%% file: main.tex
\documentclass[letterpaper,10pt,conference]{ieeeconf}

\IEEEoverridecommandlockouts
\usepackage{amsmath}
\usepackage{amssymb}
\usepackage{graphicx}
\usepackage{booktabs}
\usepackage{flushend}
\usepackage[hidelinks]{hyperref}
\usepackage{url}
\DeclareMathSizes{8.5}{8.5}{6}{5}

\newcommand{\method}{DexPolicy}

\title{\LARGE \bf
DexPolicy: Scheduled Exploration for Trajectory-Guided
Dexterous Manipulation
}

\author{
    \textbf{Haoyu Wang}$^{1*}$\quad
    \textbf{Siyuan Qian}$^{1*}$\quad
    \textbf{Yanjun Li}$^{1*}$\quad
    \textbf{Zeyu Zhang}$^{1*\dag}$\\
    \textbf{Yandong Guo}$^{2}$\quad
    \textbf{Boxin Shi}$^{1}$\quad
    \textbf{Hao Tang}$^{1\ddag}$\\
    $^1$School of Computer Science, Peking University\quad
    $^2$AI$^2$ Robotics\\
    \small $^*$Equal contribution. $^\dag$Project lead.
    $^\ddag$Corresponding author: bjdxtanghao@gmail.com
}

\begin{document}

\bstctlcite{dexpolicy_bibstyle}

\maketitle
\thispagestyle{empty}
\pagestyle{empty}

\input{sections/abstract}

\input{sections/introduction}

\input{sections/related_work}

\input{sections/background}

\input{sections/method}

\input{sections/experiments}

\input{sections/limitations}

\input{sections/conclusion}

\bibliographystyle{IEEEtran}
\bibliography{verified_refs}

\end{document}

%% file: sections/abstract.tex
\begin{abstract}
Reinforcement learning (RL) for dexterous manipulation must discover
finger--object contacts and then control the object precisely; the action
noise that serves the first goal can interfere with the second. In
trajectory-guided settings such as ViViDex, where RL refines hand--object
trajectories from human video, our baseline PPO runs end near their initial
action noise after 5M steps, motivating explicit control of exploration
scale. \method{} makes that scale an explicit function of training steps,
annealing from broad to narrow exploration while holding loss, architecture,
reward, and optimizer settings fixed. We study three policy-optimization
settings: PPO, critic-free GRPO continuation, and a flow-parameterized PPO
variant (FPO). Across five YCB objects and three training seeds,
mean deterministic Target success rises from 49.4\% to 68.1\%
(FPO), 14.1\% to 45.4\% (GRPO), and 32.0\% to 35.7\% (PPO). On a RealMan
RM75 arm with an Inspire/RH56 hand, 360 trials over three objects raise mean
Target success from 25.0\% to 85.0\% (FPO), 10.0\% to 63.3\% (GRPO), and
8.3\% to 43.3\% (PPO), with one trained model per object--method condition.
PPO component screening favors noise control over the tested optimizer contraction;
the selected PPO schedule yields higher mean Target success than linear decay
with the same endpoints on three tested objects.
Training return, deterministic Target success, and tolerance to execution
noise dissociate; schedules should therefore be judged by terminal task success
under the intended execution conditions, per task and policy-optimization setting.
Code: \url{https://github.com/AIGeeksGroup/DexPolicy}.
Website: \url{https://aigeeksgroup.github.io/DexPolicy}.
\end{abstract}

%% file: sections/introduction.tex
\section{INTRODUCTION}

Dexterous manipulation requires discovering coordinated finger contacts and
preserving them while moving an object. Action noise can help find a grasp
and destroy one, especially when reinforcement learning (RL) must adapt
imperfect reference trajectories rather than imitate them exactly.

We study this problem in the state-policy stage of ViViDex~\cite{vividex},
which refines video-derived hand--object trajectories through RL before
visual-policy distillation. Its PPO actor~\cite{ppo} learns a Gaussian action
standard deviation. In three mustard baseline runs, the logged scale ends at
0.185--0.186 after 5M steps from a nominal 0.20 initialization. This limited reduction
motivates a direct question:
can explicit control of exploration scale improve the final dexterous policy,
beyond changing the noise applied when that policy is evaluated?

\begin{figure*}[t]
    \centering
    \includegraphics[width=\textwidth]{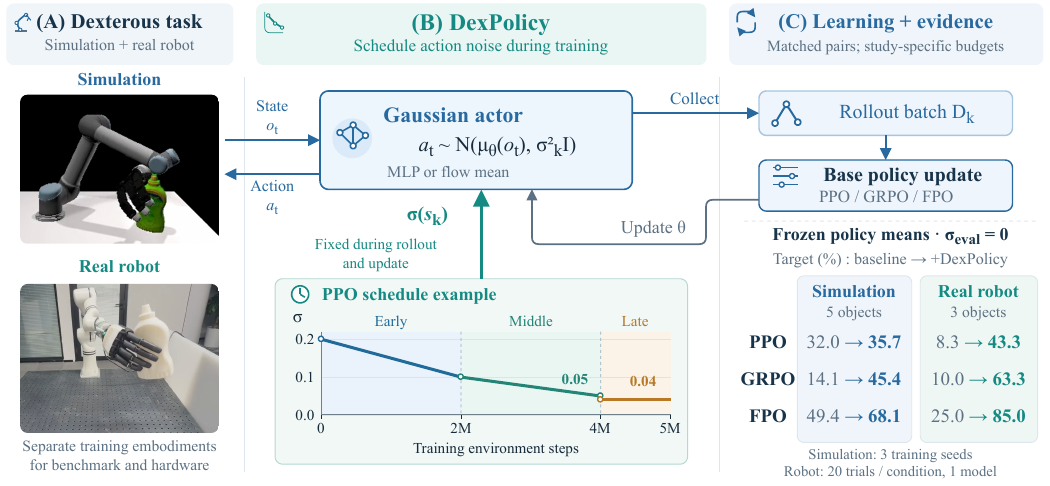}
    \caption{\method{} overview. (A) Simulation and real-robot tasks.
    (B) Gaussian noise scheduled by training steps; the PPO example does not
    represent within-trial phases or other methods' schedules.
    (C) Paired updates and deterministic Target success (\%), baseline to
    +\method{}. Simulation averages five shared objects over three seeds;
    hardware averages three objects with 20 trials from one model per condition
    (Tables~\ref{tab:full_success}--\ref{tab:real_robot}). Policies are
    object-specific, with distinct simulation and hardware embodiments;
    GRPO uses continuation budgets (Sec.~\ref{sec:grpo_continuation}).}
    \label{fig:method_overview}
\end{figure*}

\method{} schedules a scalar noise scale by training steps rather than
manipulation phase (Fig.~\ref{fig:method_overview}). Paired comparisons hold loss,
architecture, reward, and optimizer settings fixed to evaluate
final policy performance in three policy-optimization settings: PPO,
critic-free GRPO continuation, and a flow-parameterized PPO variant (FPO).

Our primary outcome is deterministic Target success: both conditions execute
their policy means ($\sigma_{\mathrm{eval}}=0$), separating learned control
from the noise applied at evaluation. We evaluate PPO and FPO after
approximately 5M training steps on eight objects, GRPO through matched
checkpoint continuations on five objects, and all three methods in
three-object hardware trials. Common-noise evaluation provides an additional
GRPO sensitivity test; native stochastic results, learning curves, controlled
ablations, and longer training provide complementary evidence.
Policies are object-specific: cross-object reuse concerns the scheduling rule,
not a transferable policy.

Our contributions are:
\begin{itemize}
    \item Scheduling exploration scale raises mean deterministic Target
    success in all three policy-optimization settings, in simulation and on
    hardware, under paired training and a common zero-noise execution rule.
    \item PPO component screening favors noise control over the tested
    optimizer contraction, and the selected PPO schedule reaches higher mean
    Target success than uniform linear decay between the same endpoints on
    the three tested objects.
    \item Training return, deterministic Target success, and tolerance to
    execution noise dissociate, so a schedule should be validated on
    terminal task success under the intended execution conditions rather
    than training return, per task and policy-optimization setting.
\end{itemize}

%% file: sections/related_work.tex
\section{RELATED WORK}

\subsection{Trajectory-Guided Dexterous Manipulation}

Human demonstrations provide useful references for dexterous control.
DexMV transfers estimated hand--object trajectories~\cite{dexmv}; DexVIP
uses human hand-pose priors for grasping~\cite{dexvip}. H-InDex studies
hand-informed visual representations~\cite{hindex}, and DexTrack learns
tracking control from human references~\cite{dextrack}. ViViDex refines
video-derived trajectories through state-based RL before visual-policy
distillation~\cite{vividex}. We retain this state-policy setting and study
exploration without adding a perception module.

\subsection{Exploration and On-Policy Optimization}

TRPO~\cite{trpo} and PPO~\cite{ppo} constrain policy updates, supporting
on-policy dexterous control~\cite{openai2020dexterous} alongside
demonstration-guided contact learning~\cite{rajeswaran2018dexterity}.
Their performance also depends on implementation
choices~\cite{andrychowicz2021onpolicy}. Generalized state-dependent
exploration structures Gaussian noise across states and time~\cite{raffin2022smooth};
colored-noise PPO studies how temporal correlation in action sampling affects
on-policy learning~\cite{hollenstein2024coloredppo}. These works address the
structure of exploration as well as its magnitude.

Hollenstein et al.\ find environment-dependent effects of Gaussian and
Ornstein--Uhlenbeck noise, including scale reduction, in off-policy
control~\cite{hollenstein2022actionnoise}. PPO-CMA modifies the objective and
separates mean and variance updates to address premature variance
contraction~\cite{hamalainen2020ppocma}. Our matched PPO-family studies examine
scalar training-step schedules, terminal success, and transfer limits without
changing update hyperparameters. GRPO uses group-relative returns instead of
a learned value baseline~\cite{shao2024deepseekmath}. Our robot-control
implementation normalizes discounted reward-to-go across complete episodes
at the same episode age. We hold this update fixed in paired continuations
to study its response to noise control.

\subsection{Flow-Parameterized Policies}

Flow matching learns continuous transformations of source
distributions~\cite{lipman2023flow}. FPO constructs a policy-ratio surrogate
from conditional flow-matching losses~\cite{mcallister2026fpo}; FPO++ develops
per-sample clipping and an asymmetric trust region for robot
control~\cite{yi2026flowcontrol}. Our FPO setting uses a conditional flow to
produce the action mean with a Gaussian likelihood (Sec.~\ref{sec:fpo}).
ReinFlow introduces learnable noise along flow trajectories for tractable
likelihood-based fine-tuning~\cite{zhang2025reinflow}. We study scheduled
output noise with a deterministic flow mean,
retaining a matched PPO-based update within each comparison.

%% file: sections/background.tex
\section{PROBLEM SETUP}

We study state-policy learning for trajectory-guided dexterous relocation.
Let $o_t$ be the state observation and $a_t\in\mathbb{R}^{d_a}$ the robot
action. At training iteration $k$, the scheduled Gaussian behavior policy is
\begin{equation}
    q_{\theta,k}(a_t\mid o_t)
    =\mathcal{N}\!\left(\mu_\theta(o_t),\sigma_k^2 I_{d_a}\right),
    \label{eq:gaussian_policy}
\end{equation}
where $\mu_\theta$ is an MLP or flow-generated mean and the scalar
$\sigma_k=\sigma(s_k)$ depends on cumulative environment steps $s_k$.
It sets a shared standard deviation across action dimensions. For each
batch $\mathcal D_k\sim q_{\theta_{\mathrm{old}},k}$, $\sigma_k$ is fixed
throughout collection and all update epochs. PPO uses the importance ratio
\begin{equation}
    \rho_t(\theta;k)=
    \frac{q_{\theta,k}(a_t\mid o_t)}
         {q_{\theta_{\mathrm{old}},k}(a_t\mid o_t)}
    \label{eq:ppo_ratio}
\end{equation}
and maximizes the clipped surrogate
\begin{equation}
\begin{split}
    \mathcal{J}_{\mathrm{PPO}}(\theta)=\mathbb{E}_t\big[\min(&
    \rho_t(\theta;k)\hat{A}_t,\\
    &\operatorname{clip}(\rho_t(\theta;k),1-\epsilon,1+\epsilon)
    \hat{A}_t)\big].
    \label{eq:ppo}
\end{split}
\end{equation}
Here $\hat A_t$ is the generalized advantage estimate~\cite{schulman2016gae}
and $\epsilon$ the clipping range.

%% file: sections/method.tex
\section{DexPolicy: SCHEDULED EXPLORATION FOR POLICY OPTIMIZATION}
\label{sec:method}

\begin{figure*}[t]
    \centering
    \includegraphics[width=0.98\textwidth]{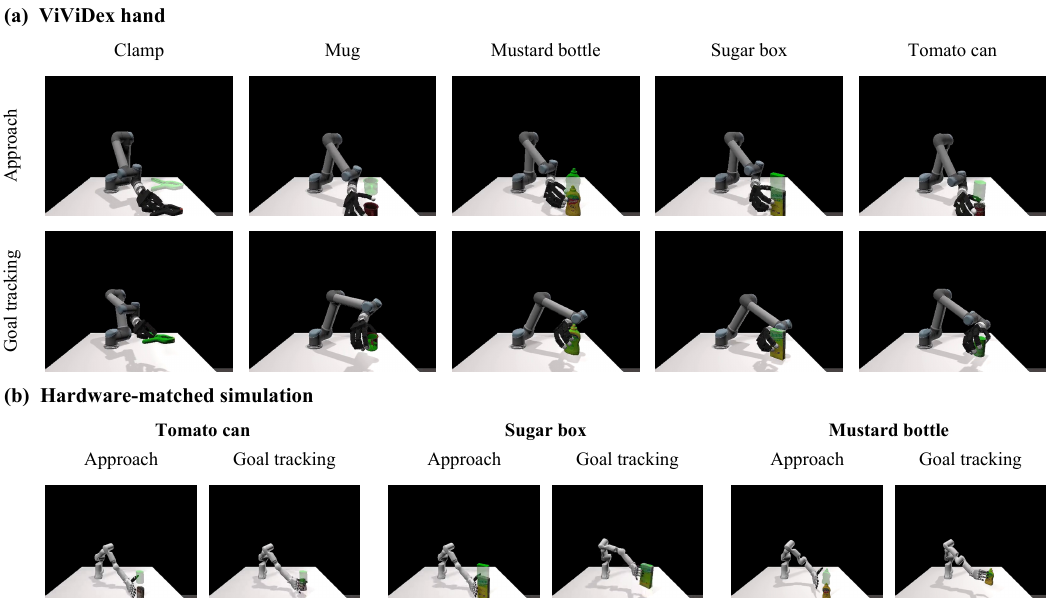}
    \caption{Simulation settings. (a) The ViViDex benchmark hand approaches
    and tracks translucent green target poses on representative objects.
    (b) Hardware-policy training uses RealMan RM75 arm and Inspire/RH56
    right-hand models matched to the real robot (Sec.~\ref{sec:real_robot}).
    These qualitative rollouts provide task context; the five-object
    comparison in Table~\ref{tab:full_success} uses banana instead of clamp.}
    \label{fig:simulation_sequences}
\end{figure*}

\method{} controls the prescribed action-noise scale over training while
preserving the loss form and optimizer settings within each paired
comparison. Changing behavior noise changes the contacts visited and the
resulting gradients; the loss form and hyperparameters do not change. Each
batch uses one standard deviation for sampling and both likelihoods in
Eq.~\eqref{eq:ppo_ratio}. The schedule follows training interactions, not
online manipulation phases: approach, grasp, and lift share the current scale.

\subsection{Proximal Policy Optimization with DexPolicy}

Baseline PPO learns log standard deviation. \method{} freezes it and
schedules the scale below, starting near the baseline initialization;
the actor mean and critic remain trainable:
\begin{equation}
\sigma(s)=
\begin{cases}
0.20-0.10\,s/(2\mathrm{M}), & 0\leq s<2\mathrm{M},\\
0.10-0.05\,(s-2\mathrm{M})/(2\mathrm{M}), & 2\mathrm{M}\leq s<4\mathrm{M},\\
0.04, & s\geq4\mathrm{M}.
\end{cases}
\label{eq:std_schedule}
\end{equation}
Learning rate $10^{-5}$, clipping range 0.20, five update epochs, and
gradient-norm cap 0.50 remain fixed. The curve is continuous at 2M and
drops from 0.05 to 0.04 at 4M.

The controlled screens motivate this curve (Sec.~\ref{sec:shape_ablation});
they do not identify a universally optimal annealing rule.

\subsection{GRPO with DexPolicy}
\label{sec:grpo_method}

Our GRPO implementation uses group-relative reward-to-go advantages for
continuous-control trajectories. For a complete trajectory $i$ of length
$T_i$, the return from step $t$ is
$G_{i,t}=\sum_{u=t}^{T_i-1}\gamma^{u-t}r_{i,u}$, with $\gamma=0.95$.
Let $\mathcal I_t=\{j:T_j>t\}$ index complete trajectories that reach
episode age $t$. For $|\mathcal I_t|\geq2$, we use
\begin{equation}
    \widehat A_{i,t}^{\mathrm{grp}}=
    \frac{G_{i,t}-\operatorname{mean}_{j\in\mathcal I_t}G_{j,t}}
         {\operatorname{std}_{j\in\mathcal I_t}G_{j,t}+10^{-8}},
    \label{eq:grpo_advantage}
\end{equation}
where the denominator uses the population standard deviation. Groups match
steps since episode start, not identical physical states. Incomplete rollout
segments and ages with fewer than two complete trajectories receive zero
advantages. The clipped objective in Eq.~\eqref{eq:ppo} uses these advantages
and averages over transitions; there is no learned value baseline, value
loss, bootstrap, or reference-policy KL penalty.

Both GRPO and GRPO+\method{} use this same update and the same starting
checkpoint within each object--seed pair. Baseline continues learning its
inherited standard deviation; \method{} freezes that parameter and applies
prescribed resets and annealing. The continuation phases and matched update
settings are specified in Sec.~\ref{sec:grpo_continuation}.

\subsection{FPO with DexPolicy}
\label{sec:fpo}

Our FPO setting replaces the MLP mean in Eq.~\eqref{eq:gaussian_policy} with a conditional
flow~\cite{lipman2023flow}. An eight-step integrator, initialized at zero,
maps normalized states to a deterministic action mean. We retain the Gaussian
likelihood and PPO clipped update. Published FPO~\cite{mcallister2026fpo} and
FPO++~\cite{yi2026flowcontrol} instead use ratios derived from flow-matching
losses. Paired flow actors and critics train from scratch with learning
rate $10^{-5}$, clip range 0.20, five epochs, and a 5M-step budget. Baseline
FPO holds $\sigma=0.10$; FPO+\method{} decays linearly from 0.20 to 0.05
over 5M steps, independently of the PPO curve in Eq.~\eqref{eq:std_schedule}.
Thus this comparison tests linear annealing against a fixed-noise flow actor;
the PPO shape screen does not select the FPO schedule.

%% file: sections/experiments.tex
\section{EXPERIMENTS}

\subsection{Benchmark and Protocol}
\label{sec:benchmark}

\input{tables/full_success}

We evaluate state-policy learning in ViViDex dexterous-hand manipulation
with YCB objects~\cite{ycb}; Fig.~\ref{fig:simulation_sequences} shows the
benchmark hand and the separate hardware-matched embodiment.
Mustard bottle is used for schedule screening; the selected curve is reused
without retuning on mug, banana, sugar box, pitcher base, and wood block.
Each object has a separately trained policy: this tests schedule transfer,
not zero-shot policy transfer. PPO comparisons hold the state observation,
22-dimensional action, reward, resets, curriculum, networks, rollout batch,
and budget fixed, with 32 training environments; learning curves average 25
stochastic episodes over five environments every 250k steps.

The FPO study uses mustard bottle, mug, sugar box, tomato can, and
extra-large clamp with three paired seeds, identical settings except
exploration, eight training and two evaluation environments, and 25 stochastic
episodes every 200k transitions. Normalized trapezoidal AUC uses logged
transitions over the common 0--5M interval.

Deterministic Target success is the principal final-policy outcome; Lift
is complementary. Both conditions execute policy means
($\sigma_{\mathrm{eval}}=0$). Target requires specified contacts and final
position within 3\,cm of the target; Lift requires specified contacts and
height above 5\,cm, which need not achieve the target position. In separate
5M learning-curve studies, normalized reward AUC measures return throughout
training; pre-grasp success, terminal contact, and lift height are diagnostics,
not completed tasks.

\subsection{Final-Policy Success: PPO and FPO}
\label{sec:full_success}

This comparison asks whether scheduling changes the policy that is finally
deployed, with sampled execution noise removed from both sides.
The eight-object comparison pairs PPO and FPO baseline/+\method{} runs over
three training seeds and a nominal 5M-transition budget, matching environment,
reward, architecture, optimizer, rollout size, and budget within pairs though
not across methods. Each checkpoint receives 100 stochastic and 100
deterministic episodes with paired evaluation seeds, one environment,
normalized trajectories, and curriculum stage 2: 96 checkpoints, 192
evaluations, and 19,200 episodes. Means and sample standard deviations
describe three training seeds, without significance claims.

Under deterministic execution, five-object mean Target success rises from
32.0\% to 35.7\% for PPO and from 49.4\% to 68.1\% for FPO
(Table~\ref{tab:full_success}). Including pitcher base, wood block, and
extra-large clamp, whose mean Target success is at most 0.34\% in every PPO/FPO execution
mode, the eight-object means are 20.0\% to 22.3\% and 30.9\% to 42.5\%.
Eight-object Lift success falls from 41.9\% to 36.0\% for PPO but rises
from 46.6\% to 50.8\% for FPO. PPO improves each eight-object mean on 1/3
paired seeds; FPO improves both on 2/3.

Table~\ref{tab:full_success} reports all three methods under paired
within-method protocols, not matched training across methods.
Aggregates hide a seed-level effect. Per seed (seeds 0/1/2, baseline to
+\method{}), scheduling removes the failed FPO seeds on mustard
(88/67/0\% to 100/100/100\%) and tomato (95/0/92\% to 95/88/95\%) but not
on mug (65/0/11\% to 0/0/93\%), which drives the large mug SD.

Eight-object mean native stochastic Target success rises from 4.5\% to 18.0\%
for PPO and 14.3\% to 38.4\% for FPO, but under unequal checkpoint noise: the
PPO baseline learns per-action standard deviations (per-model dimension means
0.185--0.192) against 0.04 with \method{}, and FPO uses 0.10 against 0.05.
These measure whole stochastic policies, not performance under common noise,
which we assess only for GRPO (Table~\ref{tab:rtg_grpo}).

\subsection{GRPO Continuation}
\label{sec:grpo_continuation}

Continuations test whether the same control transfers to an update without a
learned value baseline, resuming from trained checkpoints rather than
starting from scratch.
The GRPO continuation study comprises 30 models: five objects, two conditions,
and three seeds (Table~\ref{tab:full_success}). Each pair starts from the same
object--seed checkpoint (about 5M prior transitions recorded), sharing the
update in Sec.~\ref{sec:grpo_method}, rewards, success criteria, and 16 environments.
For the original four objects, seed 0 was used for exploratory development;
seeds 1 and 2 repeat each condition's schedule and per-object budget. These
replications reseed after checkpoint loading, whereas seed 0 retains its
earlier loading sequence; banana tests the unchanged scheme on three
reseeded, untuned runs.

Both conditions use 512 steps/environment per rollout (8,192 transitions),
minibatches of 256, five epochs, clipping range 0.20, entropy coefficient
0.001, and gradient clipping at 0.50. Two phases add 507,904 and 253,952
transitions at learning rates $10^{-4}$ and $5\times10^{-5}$. Baseline learns
its inherited standard deviation; scheduling resets it to 0.20 and anneals
$0.20\to0.10\to0.03$ across the two phases. Tomato's additional third phase
adds 253,952 transitions at $3\times10^{-5}$, with a scheduled reset to 0.10
and decay to 0.02. The intervention thus includes noise re-expansion. Budgets and learning rates
match within pairs but differ across objects.

Frozen final models receive 200 deterministic and 200 native-stochastic
episodes plus 100 at each common standard deviation, 0.03/0.10: 120
evaluations and 18,000 episodes. Evaluation matches the PPO/FPO environment,
trajectories, and Target/Lift flags at seed 25000; per-episode initial-pose
equality was not recorded.

\input{tables/rtg_grpo}

Five-object mean Target success rises from 14.1\% to 45.4\% under
deterministic execution (Table~\ref{tab:full_success}).
Banana remains difficult (3.0\% versus 3.2\% deterministic),
and mean mustard success falls from 46.25\% to 1.75\% across seeds 1 and 2
(Table~\ref{tab:rtg_grpo}). Per seed, deterministic mustard Target is
0/0/92.5\% for the baseline and 100/0.5/3.0\% with \method{} (seeds 0/1/2),
so the aggregate mustard gain is driven by seed 0. Excluding seed 0
on every object, the five-object mean still rises from 15.3\% to 37.2\%.
Deterministic Lift declines on mug (79.3\% to 76.8\%), banana (53.8\% to
42.2\%), and tomato (95.2\% to 90.7\%), while the five-object mean rises
from 78.9\% to 80.6\%. Target gains therefore need not improve lifting.

Table~\ref{tab:rtg_grpo} tests whether GRPO's deterministic gains persist under
matched execution noise. At common
$\sigma_{\mathrm{eval}}=0.03$, five-object Target is $14.8\pm8.5\%$ versus
$41.8\pm14.1\%$; at 0.10 it reverses to $16.1\pm4.6\%$ versus
$8.7\pm1.6\%$ (mean success $\pm$ sample SD in percentage points across
three seed-level object means). Neither nonzero scale was
selected as optimal. With native noise, success is 7.0\% versus 43.5\%.
Improved deterministic task completion
therefore does not establish tolerance to stronger execution perturbations.

\input{sections/real_robot}

\subsection{What Explains the Gain}
\label{sec:ablations}

Two screens ask what the improvement should be attributed to: the noise
schedule itself, or the update contraction that accompanies late low noise.
The mustard seed-0 screen separates noise scheduling from update contraction
(Table~\ref{tab:component_screen}).
The initial std curve adds a 2M $0.10\rightarrow0.08$ jump to
Eq.~\eqref{eq:std_schedule}. The update schedule jointly reduces learning rate,
clipping range, epochs, and final gradient threshold after 2M and 4M.
Noise scheduling alone raises reward AUC from 19.96 to 27.36 while improving
contact and lift. Update contraction alone lowers reward AUC to 12.35 and
lift AUC to 0.00293\,m; combining both fails to recover the baseline.
In this single-seed screen, the gain comes from noise control rather than
the tested update contraction.

\input{tables/component_screen}

\subsubsection{Schedule Shape}
\label{sec:shape_ablation}
The locked shape gate requires 95\% retention of mean reward, contact, and
lift AUC relative to the initial stagewise reference, plus reward retention
on at least two of three seeds (Table~\ref{tab:shape}). Uniform linear decay
($0.20\to0.05$ over 5M steps) and continuous front-loaded decay removing both
boundary details fail. Removing only the 2M
$0.10\rightarrow0.08$ jump passes; removing the final drop fails all three
mean-AUC criteria, with 78.2\% lift retention. We therefore remove the
intermediate jump and retain the final low-noise plateau.

\begin{table}[t]
\caption{Schedule-shape AUC retention (\%) relative to the confirmed stagewise
reference. Passing requires at least 95\% retention of all three mean AUCs
and reward retention on at least two of three seeds.}
\label{tab:shape}
\centering
\fontsize{8.5}{9.5}\selectfont
\setlength{\tabcolsep}{3pt}
\begin{tabular}{lrrrr}
\toprule
Candidate & Reward & Contact & Lift & Pass \\
\midrule
Uniform linear & 91.9 & 86.2 & 110.6 & No \\
Continuous front-loaded & 93.8 & 92.5 & 95.8 & No \\
No 2M jump & \textbf{100.3} & \textbf{95.5} & \textbf{108.3} & Yes \\
No final drop & 94.3 & 91.9 & 78.2 & No \\
\bottomrule
\end{tabular}
\end{table}

\textit{Final-policy linear control.}
A separate PPO control uses uniform linear decay from $0.20$ to $0.04$
over 5M steps, holding the remaining settings fixed. With three independent
training seeds and 100 deterministic episodes per model, Target success
(mean $\pm$ sample SD; per seed) is $83.7\pm10.3\%$ on mustard (75/95/81\%),
$29.7\pm3.5\%$ on mug (33/26/30\%), and $1.3\pm0.6\%$ on banana (1/2/1\%),
compared with 84.0\%, 25.3\%, and 8.7\% for learned-std PPO and 94.7\%,
40.7\%, and 2.3\% for \method{} (Table~\ref{tab:full_success}). Linear decay
has a similar mean to the baseline on mustard and recovers 4.3 of the
15.3-point mug gain; both schedules nearly fail on banana. Its seeds are not
paired with the benchmark seeds, and the shape result is PPO-specific:
FPO+\method{} is itself a linear schedule.

\subsection{Training-Period Outcomes and Their Limits}
\label{sec:training_period}

Return and physical proxies measured during training answer a different
question than final success, and the two need not agree.
Table~\ref{tab:screen} tests the selected PPO curve on six objects. Reward and
contact AUC improve on every object mean: reward wins 7/9 pairs on mustard,
mug, and banana, and 9/9 on the three additional geometries (16/18 overall).
Lift-height AUC improves on five objects; pitcher base declines 13.4\%.
Extra-large clamp failed the 95\% lift-retention gate on its seed-0 pair
(79.3\%) and was excluded from this transfer study.

\begin{table}[t]
\caption{Three-seed 5M schedule transfer with separately trained object policies.
Entries are relative changes (\%) in mean normalized AUC; wins count paired reward-AUC gains.}
\label{tab:screen}
\centering
\fontsize{8.5}{9.5}\selectfont
\setlength{\tabcolsep}{3pt}
\resizebox{\columnwidth}{!}{%
\begin{tabular}{@{}lrrrr@{}}
\toprule
Object & Reward AUC & Wins & Contact AUC & Lift AUC \\
\midrule
Mustard bottle & +43.4 & 3/3 & +35.4 & +38.2 \\
Mug & +14.0 & 2/3 & +27.4 & +7.6 \\
Banana & +25.1 & 2/3 & +20.1 & +55.3 \\
Sugar box & +26.1 & 3/3 & +11.3 & +144.4 \\
Pitcher base & +29.5 & 3/3 & +38.4 & $-$13.4 \\
Wood block & +64.2 & 3/3 & +240.8 & +45.2 \\
\bottomrule
\end{tabular}
}
\end{table}

\subsubsection{Longer Budget}
With unchanged PPO settings, mustard-bottle reward-AUC gains persist through
10M steps, including the 5--10M window; contact and lift AUC also improve
(Table~\ref{tab:long_budget}).

We reevaluate all six final checkpoints with 100 stochastic episodes each,
paired evaluation seeds, and no retraining. Mean reward rises from 62.30 to 69.26 (+11.2\%), with paired gains
of +6.69, +5.67, and +8.52. Contact remains 0.993; mean lift height rises from
0.1066 m to 0.1095 m (+2.8\%) with three paired wins.

\begin{table}[t]
\caption{Three-seed 10M mustard-bottle confirmation. Entries are seed means;
AUC is divided by its training window. Contact is a 0--1 fraction; lift is
height in metres. $\Delta$ is relative change; wins count paired improvements.}
\label{tab:long_budget}
\centering
\fontsize{8.5}{9.5}\selectfont
\setlength{\tabcolsep}{3pt}
\begin{tabular}{lrrrr}
\toprule
Window / metric & PPO & Scheduled & $\Delta$ (\%) & Wins \\
\midrule
Reward AUC, 0--10M & 36.57 & 47.79 & +30.7 & 3/3 \\
Reward AUC, 5--10M & 53.09 & 66.44 & +25.2 & 3/3 \\
Reward, last 20\% of steps & 59.51 & 68.28 & +14.7 & 3/3 \\
Contact AUC, 0--10M & 0.620 & 0.704 & +13.5 & 3/3 \\
Lift AUC, 0--10M (m) & 0.0654 & 0.0731 & +11.7 & 3/3 \\
\bottomrule
\end{tabular}
\end{table}

\subsubsection{Flow Action Mean}
\input{tables/flow_metrics}
Table~\ref{tab:flow_mean} complements the return histories in
Fig.~\ref{fig:learning_curves} with integrated reward, contact, and lift.
Mean final reward and reward AUC improve on all five objects, with 13/15
paired reward-AUC wins. Reward AUC rises by less than 1\% on mug and sugar
while both physical proxies decline; mustard and tomato improve all three AUCs.
Mug seed 0 also loses 31.2\% final reward.

Figure~\ref{fig:learning_curves} carries the same warning: FPO on sugar
trails before overtaking, PPO's sugar return rises while final Target success
remains zero, and GRPO's mustard return rises despite the mean Target
reversal across its replication seeds. Higher return does not guarantee
completed tasks or consistency across seeds.

%% file: tables/full_success.tex
\begin{table*}[t]
\caption{Target success: mean (\%) $\pm$ sample SD (percentage points) over three training seeds; Overall: mean over five shared objects within each seed.
Eight-object PPO/FPO means: Sec.~\ref{sec:full_success}.
PPO/FPO: 5M steps, 100 episodes/model/mode.
GRPO: ${\sim}$5M initialization + 0.762M steps (tomato: 1.016M), 200 episodes/model; seed 0 is developmental except on banana.
All methods execute policy means ($\sigma_{\mathrm{eval}}=0$).
Bold compares within pairs; budgets differ across methods.}
\label{tab:full_success}
\centering
\begingroup
\fontsize{9}{10}\selectfont
\setlength{\tabcolsep}{4pt}
\renewcommand{\arraystretch}{1.00}
\begin{tabular*}{\textwidth}{@{\extracolsep{\fill}}l*{6}{r}@{}}
\toprule
Method & \shortstack{Mustard\\bottle} & Mug & Banana & \shortstack{Sugar\\box} & \shortstack{Tomato\\can} & \shortstack{Overall\\(5 objects)} \\
\midrule
PPO baseline & 84.0\,\ensuremath{\pm}\,5.3 & 25.3\,\ensuremath{\pm}\,9.5 & \textbf{8.7\,\ensuremath{\pm}\,13.3} & 0.0\,\ensuremath{\pm}\,0.0 & \textbf{42.0\,\ensuremath{\pm}\,14.7} & 32.0\,\ensuremath{\pm}\,5.3 \\
\textbf{PPO + \method{}} & \textbf{94.7\,\ensuremath{\pm}\,9.2} & \textbf{40.7\,\ensuremath{\pm}\,17.1} & 2.3\,\ensuremath{\pm}\,2.1 & 0.0\,\ensuremath{\pm}\,0.0 & 40.7\,\ensuremath{\pm}\,9.1 & \textbf{35.7\,\ensuremath{\pm}\,3.9} \\
\addlinespace[2pt]
GRPO baseline & 30.8\,\ensuremath{\pm}\,53.4 & 25.3\,\ensuremath{\pm}\,15.0 & 3.0\,\ensuremath{\pm}\,3.9 & 4.2\,\ensuremath{\pm}\,4.3 & 7.0\,\ensuremath{\pm}\,3.1 & 14.1\,\ensuremath{\pm}\,8.5 \\
\textbf{GRPO + \method{}} & \textbf{34.5\,\ensuremath{\pm}\,56.7} & \textbf{41.7\,\ensuremath{\pm}\,3.3} & \textbf{3.2\,\ensuremath{\pm}\,2.8} & \textbf{88.0\,\ensuremath{\pm}\,12.3} & \textbf{59.8\,\ensuremath{\pm}\,4.5} & \textbf{45.4\,\ensuremath{\pm}\,14.4} \\
\addlinespace[2pt]
FPO baseline & 51.7\,\ensuremath{\pm}\,46.0 & 25.3\,\ensuremath{\pm}\,34.8 & 16.3\,\ensuremath{\pm}\,8.0 & 91.3\,\ensuremath{\pm}\,13.3 & 62.3\,\ensuremath{\pm}\,54.0 & 49.4\,\ensuremath{\pm}\,21.5 \\
\textbf{FPO + \method{}} & \textbf{100.0\,\ensuremath{\pm}\,0.0} & \textbf{31.0\,\ensuremath{\pm}\,53.7} & \textbf{17.7\,\ensuremath{\pm}\,16.6} & \textbf{99.0\,\ensuremath{\pm}\,1.7} & \textbf{92.7\,\ensuremath{\pm}\,4.0} & \textbf{68.1\,\ensuremath{\pm}\,11.4} \\
\bottomrule
\end{tabular*}
\endgroup
\end{table*}

%% file: tables/rtg_grpo.tex
\begin{table}[t]
\caption{GRPO Target change (+\method{} minus baseline), in percentage points. Seeds 1/2: replication on the original four objects; subset of untuned banana seeds. Episodes/model: 200 at zero noise, 100 otherwise.}
\label{tab:rtg_grpo}
\centering
\begingroup
\fontsize{8.5}{9.5}\selectfont
\setlength{\tabcolsep}{2pt}
\begin{tabular*}{\columnwidth}{@{\extracolsep{\fill}}lrrrrr@{}}
\toprule
Evaluation & Mustard & Mug & Banana & Sugar & Tomato \\
\midrule
Seeds 1/2, $\sigma=0$ & $-44.5$ & $+23.5$ & $+0.2$ & $+80.5$ & $+49.8$ \\
\addlinespace[2pt]
All seeds, $\sigma=0.03$ & $+6.3$ & $+13.3$ & $+0.3$ & $+72.0$ & $+43.0$ \\
All seeds, $\sigma=0.10$ & $-6.7$ & $+2.7$ & $+0.7$ & $-32.0$ & $-1.7$ \\
\bottomrule
\end{tabular*}
\endgroup
\end{table}

%% file: sections/real_robot.tex
\subsection{Real-Robot Evaluation}
\label{sec:real_robot}

\input{tables/real_robot}

\begin{figure*}[t]
    \centering
    \includegraphics[width=\textwidth]{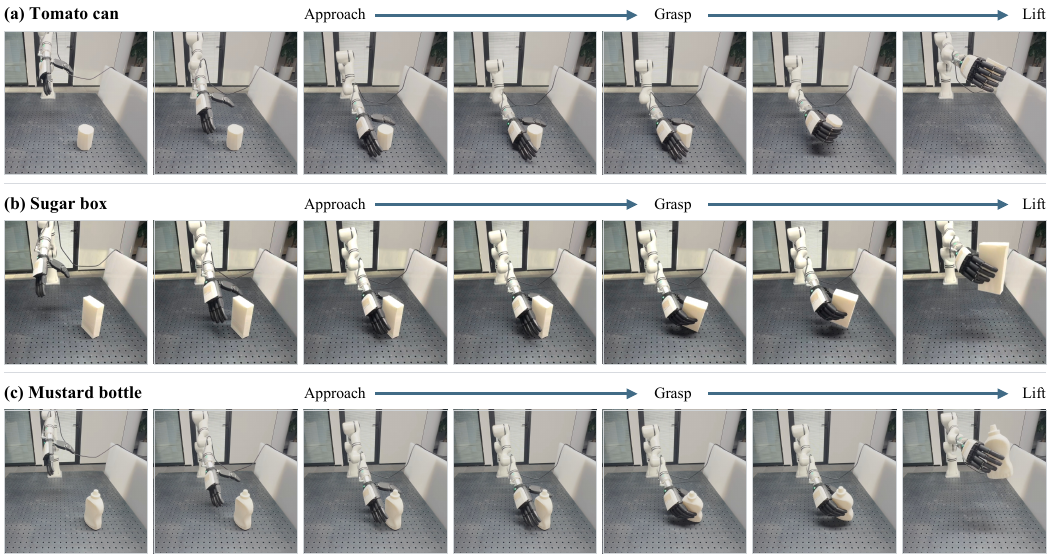}
    \caption{Grasp-and-lift sequences with a RealMan RM75 arm and an
    Inspire/RH56 right hand.
    The physical objects match the (a) tomato-can, (b) sugar-box, and
    (c) mustard-bottle geometries used in simulation. Each row shows seven
    uniformly sampled frames illustrating task progression in a demonstration
    video. These sequences illustrate the tasks rather than evaluation
    completion times.}
    \label{fig:real_robot_tasks}
\end{figure*}

\begin{figure*}[t]
    \centering
    \includegraphics[width=\textwidth]{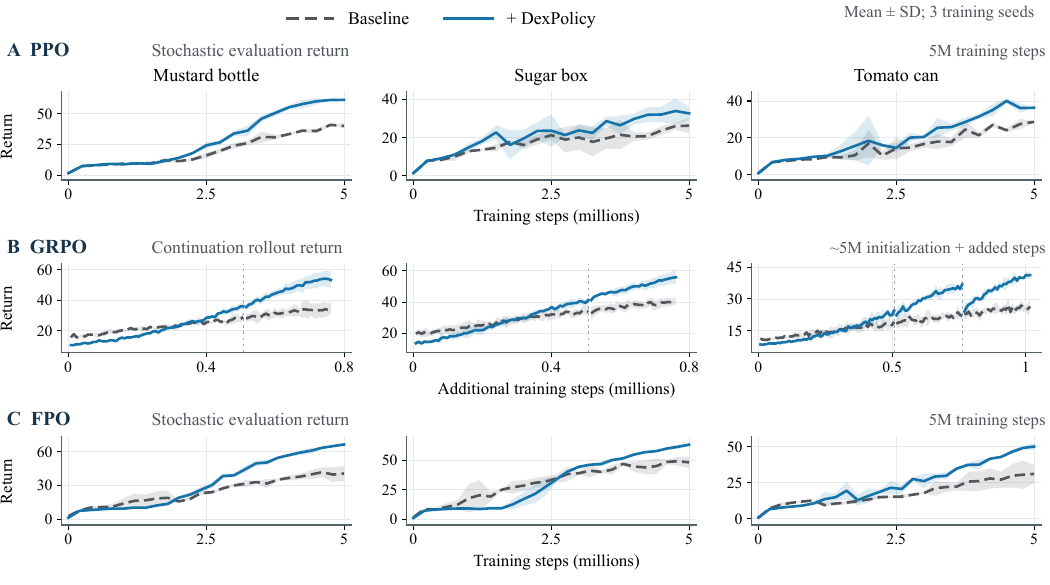}
    \caption{Learning histories on the three hardware object geometries,
    using separate benchmark policies. PPO/FPO show stochastic evaluation
    returns; GRPO shows continuation rollout returns. Shading: sample SD
    over three seeds; GRPO includes development seed 0. Dotted lines mark
    GRPO phase reloads; PPO checkpoints are nominal. Compare conditions
    within panels: return measurements and budgets differ across methods.}
    \label{fig:learning_curves}
\end{figure*}

Hardware trials test whether scheduling training noise improves
deterministic control on a physical platform.

\textit{Deployment.}
We compare PPO, GRPO, and FPO with and without \method{} on a RealMan RM75
arm with an Inspire/RH56 right hand. Policies train in the embodiment-matched
simulator in Fig.~\ref{fig:simulation_sequences}(b), separate from the ViViDex
benchmark in Fig.~\ref{fig:simulation_sequences}(a). Each condition retains its
corresponding simulation noise rule: baseline noise treatment or
\method{}'s exploration-scale control. Budgets match the corresponding
simulation experiments; baseline/\method{} training seeds and initializations
are paired. All conditions deploy the final checkpoint with frozen weights,
without physical fine-tuning or visual-policy distillation. Observations
comprise arm/hand joint states, end-effector state, and model-based RGB-D
FoundationPose~\cite{foundationpose} object poses in the robot base frame;
ordering, units, and normalization match training. At 20\,Hz, policies output
end-effector pose and active hand-joint position targets through shared inverse
kinematics and position controllers with low-level interpolation. All methods
share perception/control interfaces and execute policy means
($\sigma_{\mathrm{eval}}=0$).

\textit{Trial protocol.}
Figure~\ref{fig:real_robot_tasks} illustrates approach, grasp, and lift on
the three simulation-matched object geometries. Each of 18 object--method conditions uses
one model and 20 trials (360 total). Per object, six methods reuse 20 positions
randomly drawn within a $30\!\times\!30$\,cm planar workspace, with randomized
method order. Trials allow 20\,s from first inference to secure the object,
then 20\,s from confirmed stable grasp for lifting and verification (40\,s
maximum); autonomous regrasping does not reset either clock. Knockover,
dropping, or manual assistance ends the trial as a failure, and retries do
not replace failures. Target and Lift use the simulation thresholds of
Sec.~\ref{sec:benchmark}, scored separately on the same trials and held for
2\,s within the lifting window; stable grasp alone does not imply Target
success, and a missed target does not invalidate a successful lift.

\textit{Results.}
Mean target success rises from 8.3\% to 43.3\% for PPO, from 10.0\%
to 63.3\% for GRPO, and from 25.0\% to 85.0\% for FPO
(Table~\ref{tab:real_robot}). Mean lift success rises from 38.3\% to
58.3\% for PPO, from 46.7\% to 95.0\% for GRPO, and from 45.0\% to
93.3\% for FPO. PPO target success on sugar box remains zero. Since all
conditions execute deterministically, these gains cannot be attributed to
sampling less Gaussian noise at deployment. Twenty trials estimate each
model's performance, not variability across independent training runs.

%% file: tables/real_robot.tex
\begin{table}[t]
\caption{Deterministic real-robot success rates (\%): 20 trials per object--method condition, one model per condition. Bold marks the higher rate within each pair.}
\label{tab:real_robot}
\centering
\begingroup
\fontsize{8.5}{9.5}\selectfont
\setlength{\tabcolsep}{2pt}
\renewcommand{\arraystretch}{1.05}
\begin{tabular*}{\columnwidth}{@{\extracolsep{\fill}}lrrrrrr@{}}
\toprule
Method & \multicolumn{2}{c}{Tomato can} & \multicolumn{2}{c}{Sugar box} & \multicolumn{2}{c}{Mustard bottle} \\
\cmidrule(lr){2-3}\cmidrule(lr){4-5}\cmidrule(lr){6-7}
 & Target & Lift & Target & Lift & Target & Lift \\
\midrule
PPO baseline & 10 & 35 & 0 & 5 & 15 & 75 \\
\textbf{PPO + \method{}} & \textbf{45} & \textbf{70} & 0 & \textbf{10} & \textbf{85} & \textbf{95} \\
\addlinespace[2pt]
GRPO baseline & 5 & 35 & 10 & 75 & 15 & 30 \\
\textbf{GRPO + \method{}} & \textbf{55} & \textbf{90} & \textbf{90} & \textbf{100} & \textbf{45} & \textbf{95} \\
\addlinespace[2pt]
FPO baseline & 25 & 30 & 20 & 45 & 30 & 60 \\
\textbf{FPO + \method{}} & \textbf{80} & \textbf{95} & \textbf{75} & \textbf{85} & \textbf{100} & \textbf{100} \\
\bottomrule
\end{tabular*}
\endgroup
\end{table}

%% file: tables/component_screen.tex
\begin{table}[t]
\caption{Mustard seed-0 component screen: normalized AUC over 5M steps.
Noise uses the initial curve with the 2M jump. Contact is a fraction; lift is height.}
\label{tab:component_screen}
\centering
\begingroup
\fontsize{8.5}{9.5}\selectfont
\setlength{\tabcolsep}{2pt}
\begin{tabular*}{\columnwidth}{@{\extracolsep{\fill}}lrrr@{}}
\toprule
Condition & Reward & Contact & Lift (m) \\
\midrule
Learned-std PPO & 19.96 & 0.325 & 0.02995 \\
Noise schedule only & 27.36 & 0.427 & 0.03139 \\
Update contraction only & 12.35 & 0.200 & 0.00293 \\
Both & 19.13 & 0.217 & 0.00327 \\
\bottomrule
\end{tabular*}
\endgroup
\end{table}

%% file: tables/flow_metrics.tex
\begin{table}[t]
\caption{FPO diagnostics (three seeds): final reward mean $\pm$ sample SD
and relative AUC changes over 0--5M steps. Contact measures count
rather than PPO's fraction; lift measures height.}
\label{tab:flow_mean}
\centering
\begingroup
\fontsize{8.5}{9.5}\selectfont
\setlength{\tabcolsep}{2pt}
\begin{tabular*}{\columnwidth}{@{\extracolsep{\fill}}lrrrrr@{}}
\toprule
Object & \multicolumn{2}{c}{Final reward} & \multicolumn{3}{c}{AUC $\Delta$ (\%)} \\
\cmidrule(lr){2-3}\cmidrule(lr){4-6}
 & Baseline & Scheduled & Reward & Contact & Lift \\
\midrule
Mustard & $40.53\!\pm\!6.78$ & $66.33\!\pm\!1.09$ & $+31.3$ & $+14.2$ & $+100.6$ \\
Mug & $30.05\!\pm\!5.82$ & $34.56\!\pm\!14.64$ & $+0.7$ & $-20.3$ & $-23.0$ \\
Sugar & $48.08\!\pm\!4.79$ & $63.09\!\pm\!1.19$ & $+0.8$ & $-15.2$ & $-20.2$ \\
Tomato & $31.11\!\pm\!6.33$ & $50.15\!\pm\!2.28$ & $+38.8$ & $+47.0$ & $+142.1$ \\
Clamp & $17.62\!\pm\!1.20$ & $20.63\!\pm\!0.83$ & $+8.1$ & $+0.3$ & $+144.7$ \\
\bottomrule
\end{tabular*}
\endgroup
\end{table}

%% file: sections/limitations.tex
\section{LIMITATIONS}

Object-specific policies with one reference per object do not establish
zero-shot generalization. Three of eight PPO/FPO objects remain effectively
unsolved; extended training covers only mustard. Ablations cover one optimizer
contraction and one fixed FPO scale. Without a tuned fixed low-noise PPO
baseline, these tests do not establish superiority over equally tuned
alternatives.

The mean mustard reversal across GRPO replication seeds limits
repeatability. Continuations establish
neither from-scratch efficiency nor budget-matched cross-method performance.
Resetting, freezing, and annealing remain unseparated; the shared
credit-assignment rule is not separately ablated.

Native stochastic PPO/FPO evaluations mix policy quality with execution
noise, and matched-noise diagnostics cover only GRPO at two scales.

Three training seeds limit stability estimates. Hardware uses one model per
condition on one platform; repeated trials cannot establish training-seed
robustness or cross-platform transfer. Its simulator differs from the
benchmark, so score differences do not measure a controlled sim-to-real gap.

%% file: sections/conclusion.tex
\section{CONCLUSION}

In trajectory-guided dexterous manipulation, making exploration scale an
explicit function of training steps raises mean deterministic Target success
in three policy-optimization settings, in simulation and on hardware, without
changing the loss or the optimizer. PPO component screening favors noise
control over the tested optimizer contraction. The study also separates
three outcomes that one training curve does not distinguish:
return rises on objects whose Target success stays near zero, a Target gain
can coexist with a Lift decline, and a deterministic advantage can reverse
under stronger execution noise. Exploration scale is therefore a design
variable worth scheduling rather than leaving fixed or learned as a
by-product of the update, with its benefit measured on terminal task
success under the intended execution conditions, per task and
policy-optimization setting.